\documentclass[letterpaper, 10 pt, conference]{ieeeconf}  

\IEEEoverridecommandlockouts                              

\usepackage{graphics} 
\usepackage{epsfig} 
\usepackage{times} 
\usepackage{amsmath} 
\usepackage{amssymb}  

\usepackage{xcolor}
\usepackage{graphicx}
\usepackage{cite}
\usepackage{booktabs}

\usepackage{amsfonts}
\usepackage{todonotes}
\usepackage{subcaption}
\usepackage{algorithm}
\usepackage{algpseudocode}
\usepackage{float}
\usepackage{multirow}
\usepackage{threeparttable}
\usepackage{hyperref}
\usepackage{microtype}
\usepackage{tikz}
\usetikzlibrary{arrows.meta,positioning,fit,calc,shapes.multipart}

\title{\LARGE \bf
Chance-Constrained MPPI under State and Dynamic Object Prediction Uncertainty and the Evaluation of Collision Risk Calibration}

\author{Benjamin Serfling$^{1}$, Konrad Doll$^{1}$, and Kati Radkhah-Lens$^{1}$ 
\thanks{*This work was not supported by any organization}
\thanks{$^{1}$ All authors are with the Faculty of Engineering and Informatics, University of Applied Sciences Aschaffenburg, Aschaffenburg, Germany {\tt\footnotesize firstname.lastname@th-ab.de}}}

\begin{document}

\maketitle
\begingroup
\renewcommand\thefootnote{}
\footnotetext{This work has been submitted to the IEEE for possible publication. 
Copyright may be transferred without notice, after which this version may no longer be accessible.}
\addtocounter{footnote}{-1}
\endgroup
\thispagestyle{empty}
\pagestyle{empty}

\begin{abstract}
Chance-constrained Model Predictive Path Integral (MPPI) control is increasingly adopted for navigation in dynamic environments to explicitly bound collision risk. However, these probabilistic guarantees implicitly assume that upstream uncertainties from localization and perception are well-calibrated. In practice, estimators are often miscalibrated, inducing characteristic closed-loop failure modes: overconfidence leads to systematic safety violations, while underconfidence triggers overly conservative freezing or probability dilution. To address this critical gap, our primary contribution is a rigorous evaluation methodology applying proper scoring rules to assess the statistical validity of predicted collision risks during closed-loop execution. Concurrently, Dual-Uncertainty Chance-Constrained Tube MPPI (DUCCT-MPPI) is proposed as a real-time, risk-aware planning architecture. DUCCT-MPPI jointly integrates localization uncertainty via a one-tube Unscented Transform (UT) approximation and dynamic obstacle prediction uncertainty via Monte Carlo aggregation. Through extensive physics-based simulations, the framework demonstrates robust failure-mitigation, seamlessly transitioning to safe, conservative maneuvering without succumbing to functional deadlocks in highly cluttered environments. In highly cluttered environments, DUCCT-MPPI achieves superior robustness, outperforming established Monte Carlo MPPI baselines by nearly 28\% in navigation success rate, while simultaneously recording the lowest travel times and minimizing induced social forces. Ultimately, these findings establish that reliable probabilistic safety in autonomous navigation dictates not only expressive risk models but statistically valid uncertainty estimates throughout the entire autonomy stack.
\end{abstract}

\setlength{\abovedisplayskip}{4pt}
\setlength{\belowdisplayskip}{4pt}
\setlength{\abovedisplayshortskip}{2pt}
\setlength{\belowdisplayshortskip}{2pt}
\section{Introduction} \label{sec:Intro}

Reliable navigation in dynamic, human-populated environments requires decision-making under uncertainty from localization, perception, and prediction. Probabilistic motion planning and control methods are therefore increasingly used. In particular, Model Predictive Path Integral (MPPI) control \cite{Williams_2017_ModelPredictive,Williams_2018_InformationTheoretic} enables real-time planning in non-linear and non-convex settings, and is often paired with chance constraints to explicitly bound collision probabilities \cite{DuToit_2011_Probabilistic,Paiola_2024_Evaluation}.

These probabilistic guarantees implicitly assume that the uncertainties supplied to the planner are well-calibrated, i.e., predicted probabilities (e.g., collision risk) match empirical collision frequencies during execution. 
In practice, however, learning-based predictors are often miscalibrated \cite{Guo_2017_Calibration} and classical estimators can misjudge covariance due to modeling errors \cite{Tsuei_2023_Uncertainty}, causing a gap between planned and empirically observed risk.

This miscalibration induces characteristic closed-loop failure modes. While overconfidence can produce trajectories that formally satisfy chance constraints yet violate safety in execution, underconfidence yields overly conservative behavior such as the freezing robot phenomenon \cite{Trautman_2015_Robot,Trautman_2010_Unfreezing}. Furthermore, extreme underconfidence can cause \emph{probability dilution} via long-horizon or multi-agent marginalization. This disperses collision probability mass until local risk drops below decision thresholds, paradoxically encouraging unsafe behavior \cite{Balch_2019_Satellite,Lee_2025_Overcoming}.

Despite their relevance, such effects are rarely analyzed explicitly in chance-constrained planning, where evaluations typically emphasize collision rates or task success and treat uncertainty as ground truth. By contrast, calibration is routinely assessed with proper scoring rules in statistics, state estimation, and machine learning \cite{Guo_2017_Calibration, Poczos_2012_Nonparametric}, but is seldom applied to motion planners in closed-loop navigation.

Motivated by this gap, we study how miscalibrated uncertainty affects the validity of collision risk estimates produced by chance-constrained MPPI and how this propagates into closed-loop behavior. We introduce Dual-Uncertainty Chance-Constrained Tube MPPI (DUCCT-MPPI), which combines a one-tube Unscented Transform (UT) approximation for localization uncertainty propagation \cite{Mohamed_2025_Towards} with Monte Carlo aggregation for dynamic obstacle predictions \cite{Trevisan_2025_Dynamic}. The approach serves both as a navigation algorithm and an evaluation instrument, enabling controlled calibration manipulation and systematic comparison of predicted risks against empirical outcomes in physics-based simulations.
\section{RELATED WORK} \label{sec:related_work}

\subsection{Risk-Aware MPPI, Chance Constraints, and Uncertainty Propagation}
MPPI is widely adopted for real-time navigation due to its gradient-free formulation and GPU-parallelizable sampling \cite{Williams_2017_ModelPredictive, Williams_2018_InformationTheoretic}. Probabilistic safety is commonly enforced via chance constraints (CC) that bound collision probability by a threshold $\sigma$ \cite{DuToit_2011_Probabilistic, Zhu_2019_Chance}. Since exact collision probability evaluation is generally intractable, existing CC-MPPI approaches rely on approximations, including analytic reformulations under Gaussian assumptions \cite{Zhu_2019_Chance, Mohamed_2025_Chance} and scenario-based or sampling-based approximations for non-Gaussian uncertainty \cite{degroot_2023_scenariobasedmotionplanningbounded, Brito_2020_Contouring}.  

A central challenge in risk-aware MPPI is accounting for uncertainty from \emph{multiple sources}. In dynamic human environments, collision risk arises both from stochastic obstacle motion and from uncertainty in the robot’s own state due to localization and motion noise. Many CC-MPPI formulations simplify this problem by treating the robot state deterministically and focusing on obstacle uncertainty \cite{degroot_2023_scenariobasedmotionplanningbounded, Trevisan_2025_Dynamic}. To mitigate the underestimation of risk in crowded scenes, \cite{Trevisan_2025_Dynamic} approximates the joint collision probability across multiple dynamic obstacles using Monte Carlo integration, enabling real-time handling of non-Gaussian and multimodal predictions. However, localization uncertainty is not propagated into the collision probability estimate.  

Conversely, Unscented-Transform-based MPPI methods explicitly propagate robot state uncertainty through nonlinear dynamics to avoid overconfident trajectory evaluation under localization errors \cite{Mohamed_2025_Chance, Mohamed_2025_Towards}. These approaches integrate state covariance via sigma-point propagation and analytic chance constraints, but do not explicitly estimate \emph{joint} collision probabilities across multiple dynamic agents and instead rely on Gaussian and per-obstacle formulations.

Despite these parallel advancements, a critical gap remains: to date, there is no chance-constrained MPPI method that simultaneously integrates joint collision probability estimation for multimodal dynamic environments with the explicit propagation of robot localization uncertainty in a single planning framework.

\subsection{The Calibration Gap and Its Consequences}
Multiple studies report a systematic mismatch between predicted and empirically observed risk in chance-constrained planning, with observed collision frequencies often substantially lower than the theoretical CC bound, leading to overly conservative behavior \cite{Trevisan_2025_Dynamic, Zhu_2019_Chance, degroot_2023_scenariobasedmotionplanningbounded, wang_2020_nongaussianchanceconstrainedtrajectoryplanning}. Heuristic remedies, such as posterior risk assessment or parallel execution with varying risk thresholds, aim to reduce conservatism by selecting empirically safe plans \cite{mustafa_2023_probabilisticriskassessmentchanceconstrained}. However, these strategies implicitly assume that the collision probability estimates produced by the planner are statistically meaningful, and treat miscalibration primarily as a tuning or selection issue rather than an object of evaluation.  

Miscalibration induces severe closed-loop failures: overconfidence violates safety constraints, while underconfidence triggers the freezing robot phenomenon \cite{Trautman_2010_Unfreezing} or probability dilution, which paradoxically encourages unsafe behavior by dispersing risk below decision thresholds \cite{Balch_2019_Satellite, Lee_2025_Overcoming}.
Although calibration assessment via proper scoring rules and consistency tests (e.g., Brier score and divergence-based measures) is standard in statistics, state estimation, and machine learning \cite{Guo_2017_Calibration, Poczos_2012_Nonparametric}, it remains uncommon for motion planners operating in closed-loop navigation.  

To date, the literature lacks a systematic evaluation of whether theoretical collision probabilities match empirical collision rates in closed-loop execution. In this work, we address this critical gap and explicitly study how such miscalibration affects navigation behavior.

\subsection{Contributions}

While the proposed planner synthesizes established techniques in uncertainty propagation, its primary purpose is to serve as an enabling instrument for our core contribution: the rigorous closed-loop analysis of risk calibration. Specifically, our contributions are threefold:
\begin{itemize}
    \item \textbf{Calibration-Focused Evaluation Methodology:} We introduce the application of proper scoring rules (e.g., Brier score, Log Loss) to chance-constrained motion planning, providing a rigorous framework to evaluate the statistical validity of predicted collision risks.
    \item \textbf{Identification of Miscalibration Failure Modes:} We empirically demonstrate how localization and prediction miscalibration induce systematic failure modes: overconfidence causes safety violations, while underconfidence triggers freezing or probability dilution.
    \item \textbf{DUCCT-MPPI Architecture:} We formulate a computationally tractable planning architecture integrating the Unscented Transform (UT) with Monte Carlo dynamic obstacle aggregation, specifically designed to expose and evaluate these closed-loop calibration effects.
\end{itemize}
\section{Preliminaries}

We consider closed-loop navigation under uncertainty from two upstream sources: (i) \emph{robot state} uncertainty from localization and (ii) \emph{dynamic obstacle} uncertainty from prediction. As discussed in sections \ref{sec:Intro}--\ref{sec:related_work}, chance-constrained planners rely on these probabilistic inputs to be statistically meaningful. Our formulation makes these inputs explicit.

\subsection{Robot dynamics and localization belief}

The robot follows nonlinear discrete-time dynamics
\begin{equation}
\mathbf{x}_{t+1}=F(\mathbf{x}_t,\mathbf{u}_t),
\end{equation}
with state $\mathbf{x}_t=[x_t,y_t,\psi_t]^\top\in\mathbb{R}^{3}$, $\mathbf{p}_t=[x_t,y_t]^\top$ as the positional subset of this state and the control $\mathbf{u}_t\in\mathbb{R}^{n_u}$. Here $t\in\mathbb{N}$ denotes a dimensionless discrete-time index with fixed sampling period $\Delta t$.
Localization is represented as a Gaussian belief

\begin{equation}
\mathbf{x}_t\sim\mathcal{N}(\hat{\mathbf{x}}_t,\mathbf{\Sigma}_t),
\label{eq:gaussian_state}
\end{equation} 
with mean state $\hat{\mathbf{x}}_t$ and covariance matrix $\mathbf{\Sigma}_t\in\mathbb{R}^{3\times 3}$. We later use the positional block $\mathbf{\Sigma}_{p,t}\in\mathbb{R}^{2\times 2}$, where all correlations between position and orientation encoded in
$\mathbf{\Sigma}_{t}$ are discarded.

\subsection{Dynamic obstacle prediction}
Predictions of an obstacle $o$ are provided as normalized spatial occupancy probabilities
$\mathcal{P}_{o,t}(x,y)\in[0,1]$, potentially multi-modal (e.g., mixtures). We assume conditional independence across obstacles given the predictor output.

\subsection{Chance-constrained objective}
We seek a control sequence $U=\{\mathbf{u}_0,\dots,\mathbf{u}_{T-1}\}$  for a prediction horizon of $T$ discrete time steps (physical duration $T\Delta t$), minimizing the expected finite-horizon cost
\begin{equation}
\min_{U} \quad \mathbb{E}\!\left[\phi(\mathbf{x}_T)+\sum_{t=0}^{T-1}q(\mathbf{x}_t,\mathbf{u}_t)\right],
\label{eq:MPPI_cost}
\end{equation}

where $\phi(\mathbf{x}_T)$ denotes a terminal cost and $q(\mathbf{x}_t,\mathbf{u}_t)$ a stage cost.
The optimization is subject to a time-indexed joint collision chance constraint across all obstacles:
\begin{equation}
\mathbb{P}(C_t)\le 1-\sigma,\qquad \forall t,
\end{equation}
where $C_t$ denotes the event that the robot footprint intersects any obstacle at time $t$, and $\sigma\in ]0,1[$ is the constant chance constraint, which specifies the maximum allowable collision probability.

\subsection{MPPI optimization}
We solve the resulting nonconvex stochastic optimization problem using MPPI \cite{Williams_2018_InformationTheoretic}.
At each time cycle, $K$ control sequences $\{U_k\}_{k=0}^{K-1}$ with
$U_k=\{\mathbf{u}_{k,0},\dots,\mathbf{u}_{k,T-1}\}$ are sampled by perturbing a nominal warm-start sequence
$U_{\text{init}}=\{\mathbf{u}_0,\dots,\mathbf{u}_{T-1}\}$ with Gaussian noise:
\begin{equation}
U_k = U_{\text{init}} + \mathcal{E}_k,\qquad
\mathcal{E}_k=\{\boldsymbol{\epsilon}_{k,t}\sim\mathcal{N}(\mathbf{0},\mathbf{\Sigma}_u)\}_{t=0}^{T-1}.
\end{equation}
Each rollout $U_k $ induces a state trajectory $\{\mathbf{x}_{k,t}\}_{t=0}^{T}$ and a corresponding trajectory cost $S_k$, obtained by evaluating \eqref{eq:MPPI_cost} along the $k$-th sampled trajectory, including collision-risk penalties implied by the chance constraint.
\begin{equation}
w_k=\frac{1}{\eta}\exp\!\left(-\frac{1}{\beta}(S_k-\rho)\right),\qquad
U^*=\sum_{k=0}^{K-1} w_k U_k,
\label{eq:mppi_update}
\end{equation}
with $\rho=\min_k (S_k)$ for numerical stability and $\eta=\sum_k \exp(-\tfrac{1}{\beta}(S_k-\rho))$ \cite{Trevisan_2025_Dynamic}.
Only the first control input $\mathbf{u}^*_0$ of $U^*$ is applied to the system. $U^*$ is the cost-weighted mean of sampled control sequences, approximating the minimizer of the stochastic optimal control problem. The sequence $U^*$ is shifted by one timestep to warm-start the next cycle.

\section{Proposed DUCCT-MPPI Method}

We propose \textbf{D}ual-\textbf{U}ncertainty \textbf{C}hance-\textbf{C}onstrained \textbf{T}ube MPPI (DUCCT),
a one-tube, uncertainty-aware MPPI method that (i) propagates localization uncertainty once per cycle and
(ii) evaluates joint collision risk by combining analytic robot occupancy with Monte Carlo aggregation of obstacle occupancies.

To make the integration of multiple uncertainty sources computationally tractable for real-time control, DUCCT is organized as a sequential pipeline:
\begin{enumerate}
    \item \textbf{One-Tube Propagation (Sec. \ref{subsec:ducct_one_tbe}):} Robot localization uncertainty is propagated once per cycle along a nominal trajectory.
    \item \textbf{Analytic Occupancy (Sec. \ref{subsec:ducct_occupancy}):} This positional variance is converted into an analytic spatial footprint.
    \item \textbf{Joint Risk Aggregation (Sec. \ref{subsec:ducct_monte_carlo_joint} \& \ref{sec:predictor_model}):} The robot's footprint is evaluated against predicted dynamic obstacles using Monte Carlo sampling.
    \item \textbf{MPPI Integration (Sec. \ref{subsec:ducct_risk_integrattion}):} The resulting collision probabilities are applied as both soft costs and hard chance constraints.
\end{enumerate}
For readability, we omit rollout and time indices ($k, t$) where unambiguous.

\subsection{One-Tube Uncertainty Propagation (Localization)}
\label{subsec:ducct_one_tbe}

To account for localization error, the robot state from \eqref{eq:gaussian_state} is propagated through the nonlinear dynamics using a prediction-only Unscented Transform (UT) \cite{Mohamed_2025_Towards,Mohamed_2025_Chance}. 
However, propagating belief for all $K$ rollouts is computationally prohibitive.
DUCCT uses a \emph{one-tube} approximation, where the covariance sequence $\{\mathbf{\Sigma}_t\}$ is propagated once per planning cycle along the nominal mean trajectory:
\begin{equation}
\mathbf{x}_{k,t}\sim\mathcal{N}(\hat{\mathbf{x}}_{k,t},\mathbf{\Sigma}_t),\qquad t=0,\dots,T
\end{equation}
Here, $\mathbf{x}_T$ denotes the terminal state obtained after applying the control inputs
$\{\mathbf{u}_t\}_{t=0}^{T-1}$.
To decouple belief propagation from the current MPPI sampling loop, we propagate using the previous-cycle nominal controls
$U_{prev}^*$, and we do not add process noise. Thus, $\mathbf{\Sigma}_t$ captures the propagation of the initial localization uncertainty through the nonlinear dynamics under $U_{prev}^*$. All rollouts share $\{\mathbf{\Sigma}_t\}$ while maintaining distinct mean trajectories, which is justified by the small MPPI perturbations.

\subsection{Analytic Robot Occupancy via Belief--Footprint Integration}
\label{subsec:ducct_occupancy}

With the state covariance $\{\mathbf{\Sigma}_t\}$ computed, we must translate this statistical uncertainty into a spatial collision footprint. We compute a continuous robot occupancy field $\mathcal{P}_{\text{occ}}(x,y)$ that quantifies the probability that a spatial location $(x,y)$ is covered by the robot body.
The robot footprint is approximated as a square region
$R^\square_{\text{combined}}$ centered at the rollout mean position $(x_t,y_t)$
with side length $l_{\text{combined}} = l_{\text{robot}} + l_{\text{person}}$, where $l_{\text{robot}}$ and $l_{\text{person}}$ denote the effective footprint radii of the robot and a pedestrian, respectively.
For notational convenience, we define $L = \tfrac{l_{\text{combined}}}{2}$.
The square footprint is adopted to obtain a separable collision region that
admits a closed-form evaluation of the belief--footprint integral under a
Gaussian position belief.
This enables an efficient analytic expression based on error functions and
avoids numerical integration or sampling within the MPPI loop.
The resulting footprint provides a conservative over-approximation of the
collision region suitable for risk-aware planning.

Let $\mathbf{p}_t = [x_t, y_t]^\top \sim \mathcal{N}(\hat{\mathbf{p}}_t, \mathbf{\Sigma}_{p,t})$
denote the uncertain robot position at time $t$.
We define the robot occupancy probability of a point (x,y) as
\begin{equation}
\mathcal{P}_{\text{occ}}(x,y)
\;:=\;
\mathbb{P}\!\left(
|x - x_t| \le L,\;
|y - y_t| \le L
\right).
\label{eq:occ_def}
\end{equation}

\noindent
To evaluate \eqref{eq:occ_def} in closed form, we exploit the Gaussian
structure of the belief. Since $\mathbf{\Sigma}_{p,t}$ may be correlated, we
transform the coordinates into the principal frame of the covariance.
Let
\[
\mathbf{\Sigma}_{p,t}
=
\mathbf{R}\,\mathrm{diag}(\lambda_1,\lambda_2)\mathbf{R}^\top
\]
denote the eigendecomposition of the positional covariance, and define
\[
\boldsymbol{\xi}
=
\mathbf{R}^\top
\big([x,y]^\top - \hat{\mathbf{p}}_t\big)
\]
as the query point expressed in this principal frame.
Under this transformation, the occupancy probability in
\eqref{eq:occ_def} allows the following closed-form expression:
\begin{equation}
\mathcal{P}_{\text{occ}}(x,y)
=
\frac{1}{4}
\prod_{i=1}^{2}
\left[
\mathrm{erf}\!\left(\frac{\xi_i + L}{\sqrt{2\lambda_i}}\right)
-
\mathrm{erf}\!\left(\frac{\xi_i - L}{\sqrt{2\lambda_i}}\right)
\right],
\label{eq:occ_closed_form}
\end{equation}
\noindent
where $\mathrm{erf}(\cdot)$ denotes the Gaussian error function.
Notably, $\int_{\mathbb{R}^2} \mathcal{P}_{\text{occ}}(x,y)\,dx\,dy
= l_{\text{combined}}^2$, corresponding to the footprint area rather than a
normalized probability density.

\subsection{Monte Carlo Aggregation of Joint Collision Probability}
\label{subsec:ducct_monte_carlo_joint}

Having defined the robot's spatial occupancy analytically, we now evaluate its overlap with uncertain dynamic obstacles. Obstacle trajectory distributions are provided by a goal-directed predictor (detailed in Sec. \ref{sec:predictor_model}), yielding a Gaussian belief tube $\mathcal{P}_{o,t}$ for each obstacle $o$.

To compute the joint risk efficiently, we employ Monte Carlo (MC) integration \cite{Trevisan_2025_Dynamic}.
At each time step $t$, we draw $N_{\text{MC}}$ samples $\{(x_j,y_j)\}_{j=1}^{N_{\text{MC}}}$
\begin{equation}
R^{\square}_{\text{area},t}
=
\{(x,y)\mid x_{\text{lower,t}}\le x\le x_{\text{upper,t}},\; y_{\text{lower,t}}\le y\le y_{\text{upper,t}}\},
\end{equation}
computed from rollout extrema and inflated by $L$:
\begin{align}
x_{\text{lower,t}} &= \min_k(x_{k,t}) - L, &
x_{\text{upper,t}} &= \max_k(x_{k,t}) + L, \\
y_{\text{lower,t}} &= \min_k(y_{k,t}) - L, &
y_{\text{upper,t}} &= \max_k(y_{k,t}) + L.
\end{align}
This yields one sample set per time step, reused across all $K$ rollouts.

Assuming conditional independence across the $N_o$ predicted obstacles, indexed by $o\in\{1,\dots,N_o\}$, the \emph{joint} obstacle occupancy at sample $(x_j,y_j)$ is

\begin{equation}
\mathcal{P}_{\text{joint},j,t}
=
1-\prod_{o=1}^{N_o}\left(1-\mathcal{P}_{o,t}(x_j,y_j)\right).
\label{eq:joint_occ}
\end{equation}

Building upon the MC integration idea in \cite{Trevisan_2025_Dynamic}, our contribution is to weight each obstacle sample by the analytically derived robot occupancy.
The joint collision probability for a rollout mean position at time $t$ is estimated as
\begin{equation}
\hat{\mathcal{P}}_{k,t}
\approx
\frac{l_{\text{combined}}^2}{N_{\text{MC}}}
\sum_{j=1}^{N_{\text{MC}}}
\mathcal{P}_{\text{joint},j,t}\,\mathcal{P}_{\text{occ}}(x_j,y_j),
\label{eq:collision_est}
\end{equation}
where $\mathcal{P}_{\text{occ}}(x_j,y_j)$ is evaluated using \eqref{eq:occ_closed_form}.
Compared to binary collision indicators, \eqref{eq:collision_est} yields a smooth risk estimate that accounts for correlated localization uncertainty.

\subsection{Risk Integration into MPPI (Soft + Hard)}
\label{subsec:ducct_risk_integrattion}

Finally, with $\hat{\mathcal{P}}_{k,t}$ computed for every rollout, we integrate this risk into the MPPI optimizer as both a soft penalty and a hard chance constraint. Each rollout cost $S_k$ is augmented by the predicted joint collision probability:
\begin{equation}
S_k \leftarrow S_k + \lambda_{\text{risk}}\sum_{t=0}^{T-1}\hat{\mathcal{P}}_{k,t},
\label{eq:risk_penalty}
\end{equation}
where $\hat{\mathcal{P}}_{k,t}$ is computed via \eqref{eq:collision_est} and where $\lambda_{\text{risk}}>0$ weights the risk penalty.
In addition, we reject rollouts that violate the hard threshold:
\begin{equation}
\hat{\mathcal{P}}_{k,t}>\sigma \;\Rightarrow\; \text{reject rollout }k.
\end{equation}

\vspace{0.25em}
Overall, DUCCT enables real-time, uncertainty-aware MPPI by (i) amortizing localization uncertainty propagation into a single tube and
(ii) computing joint collision risk through an analytic belief--footprint integral coupled with weighted MC aggregation of obstacle occupancy.

\subsection{Pedestrian Prediction Model}
\label{sec:predictor_model}
To supply the obstacle distributions $\mathcal{P}_{o,t}$ required, for our experiments pedestrian trajectories are generated by a goal-directed predictor. The mean position is advanced toward a waypoint sequence at a constant observed speed, while uncertainty is propagated via a Constant Velocity (CV) model ($\Sigma_{t+1} = F \Sigma_t F^\top + Q$). This yields a Gaussian belief tube that aligns with the intended path and diffuses naturally over the horizon.
Pedestrian trajectory distributions are generated by a goal-directed predictor that decouples mean propagation from uncertainty propagation.
The mean position is advanced toward a waypoint sequence $\mathcal{G}=\{g_1,g_2,\dots\}$ at constant speed $v$ (set to the observed velocity magnitude) and switches from $g_i$ to $g_{i+1}$ once the predicted position enters a radius $r_{\mathrm{goal}}$.

Uncertainty is represented as a Gaussian belief $\mathcal{N}(\mu_t,\Sigma_t)$.
While $\mu_t$ follows the goal-directed mean propagation, covariance is propagated with a Constant Velocity (CV) model:
\begin{equation}
\Sigma_{t+1} = F \Sigma_t F^\top + Q,
\end{equation}
where $F$ is the CV transition matrix and $Q$ is process noise.
This yields a Gaussian belief tube aligned with the intended path and increasing over the horizon due to diffusion.


\section{Experiments}



Our experimental evaluation is designed to answer two critical questions. First, how does DUCCT-MPPI scale in increasingly crowded environments compared to standard baselines (Section \ref{subsec:exp_baseline_comparison})? And crucially, how does the calibration of upstream uncertainty dictate the statistical validity and physical safety (Section \ref{subsec:exp_risk_validity}) of chance-constrained planning?
%

\subsection{System and Scenarios}  \label{subsec:exp_system_scenarios}

\begin{figure}[!htbp]
    \centering
    \includegraphics[angle=-90,width=0.5\textwidth]{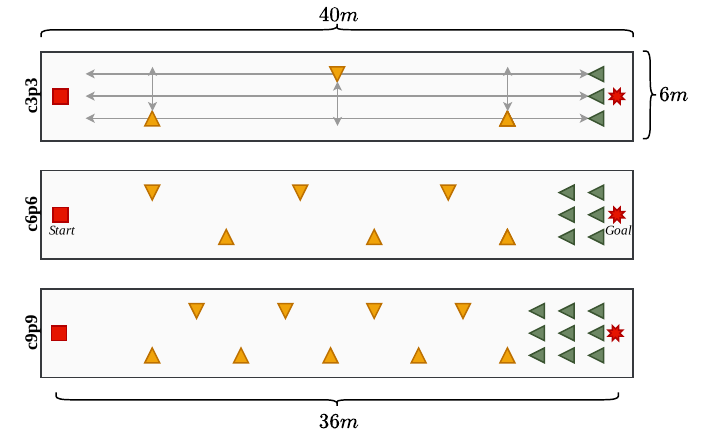}
    \caption{Scenes used for testing. Shown in red are the robot start (square) and goal positions (star). Each Scenario consists of pedestrians in passing (green) or crossing (orange) maneuvers. The movements for these maneuvers are shown in the c3p3 scenario in the upper diagram; for clarity, they are omitted from the two lower diagrams.}
    \label{fig:scenarios}
\end{figure}

Experiments are executed using the Nav2 stack \cite{macenski_2020_nav2marathon2} in Gazebo Classic simulations \cite{koenig_2004_gazebodesign} using the ODE physics engine \cite{smith_ode_2008}.
We use a TurtleBot3 Waffle Pi (differential-drive) with translational velocity limit $v_{\max}=1.0\,\mathrm{m/s}$.
Simulation and control run on a desktop PC equipped with an NVIDIA GeForce RTX~4090.
All runs take place in a $6\mathrm{m}\times 40\,\mathrm{m}$ featureless corridor, as illustrated in \autoref{fig:scenarios}.
The robot travels $36 \mathrm{m}$ by traversing the dynamic pedestrian environment.
Due to the absence of distinctive landmarks, localization is dominated by odometry drift, leading to comparatively high pose uncertainty and highlighting the benefit of explicitly accounting for localization uncertainty.
Pedestrian motion is generated in HuNavSim \cite{Noack_2022_HuNavSim} using the Social Force Model (SFM) \cite{Helbing_1995_SFM}. Although SFM is deterministic, it induces effective uncertainty for the planner, which lacks access to the true underlying motion model. We evaluate the \textit{impassive} behavior type, where pedestrians react to the robot as they would to any other object. Pedestrians are simulated with nominal walking speed $1.0\,\mathrm{m/s}$ and maximum speed $v_{\max}=1.5\,\mathrm{m/s}$.
Scenarios combine \textit{passing} and \textit{crossing} patterns \cite{gao_2022_evaluation} under three density levels as shown in \autoref{fig:scenarios}


\paragraph*{Rationale for Controlled Synthetic Evaluation}
While real-world miscalibration stems from complex perception failures, hardware or dataset evaluations introduce confounding variables that mask core algorithmic behaviors. To isolate these root causes, we deliberately employ a deterministic simulation with explicitly controlled synthetic miscalibration. This methodology acts as a rigorous in vitro stress test, decoupling the planner's response to uncertainty from the structural errors of specific perception stacks.

\subsection{Baseline Comparison} \label{subsec:exp_baseline_comparison}
\subsubsection{Controllers} \label{subsubsec:exp_controllers}
We compare three MPPI-based Nav2 local planners:
\textbf{Vanilla MPPI}, a Nav2 native MPPI controller based on \cite{Williams_2016_Aggressive},
\textbf{DRA-MPPI}, which computes joint collision probability via Monte Carlo sampling \cite{Trevisan_2025_Dynamic},
and our proposed approach \textbf{DUCCT-MPPI} integrating localization and prediction uncertainty with GPU-accelerated Monte Carlo risk evaluation.
All controllers run as Nav2 controller plugins at $20\,\mathrm{Hz}$.
Robot localization is provided by Adaptive Monte Carlo Localization (AMCL) \cite{fox_1999_monte}, and pedestrians are tracked and predicted by the probabilistic predictor from \autoref{sec:predictor_model}.
The predictor runs at $20\,\mathrm{Hz}$ with $2\,\mathrm{s}$ prediction horizon, i.e., $40$ discrete steps, matching the MPPI update frequency used by all controllers.

\subsubsection{Controller Parametrization} \label{subsubsec:exp_param}
To ensure a fair comparison, all controllers use identical MPPI parameters:
planning horizon $T=40$ with $\Delta t=0.05\,\mathrm{s}$ ($2.0\,\mathrm{s}$ horizon), $K=400$ rollouts, and a single MPPI iteration per control cycle. Trajectory weighting uses temperature $\beta=0.25$.
Risk-aware methods (DRA-MPPI, DUCCT-MPPI) approximate collision probabilities with Monte Carlo integration using $N_{\mathrm{MC}}=20{,}000$ samples per time step.
The risk threshold is set to $\sigma = 0.95$, aligning with the standard $2\sigma$ confidence interval to provide a rigorous, dynamically feasible upper bound.
Robot and pedestrians are modeled as circular footprints with radii $l_{\text{robot}}=0.3\,\mathrm{m}$ and $l_{\text{person}}=0.5\,\mathrm{m}$, used consistently for planning and ground-truth collision labeling.
Monte Carlo risk evaluation is executed on the GPU using a CUDA implementation.


\subsubsection{Metrics} \label{subsubsec:exp_nav_metrics}
To evaluate navigation performance using the HuNavSim evaluator \cite{Noack_2022_HuNavSim}, we report the Success Rate (SR) of reaching a $2 \mathrm{m}$ goal region, the total Task Duration (TD), and the Average Velocity (AV). Furthermore, we quantify safety and social compliance via the Collision Rate (CR) (the percentage of time steps where the robot-human distance falls below $l_{\text{combined}}=l_{\text{robot}}+l_{\text{person}}$) and the cumulative Social Force (SF) induced by the robot on human agents \cite{Noack_2022_HuNavSim}.

\subsubsection{Results} \label{subsubsec:exp_baseline_results}
\begin{figure*}[htbp]
    \centering
    \includegraphics[width=\textwidth]{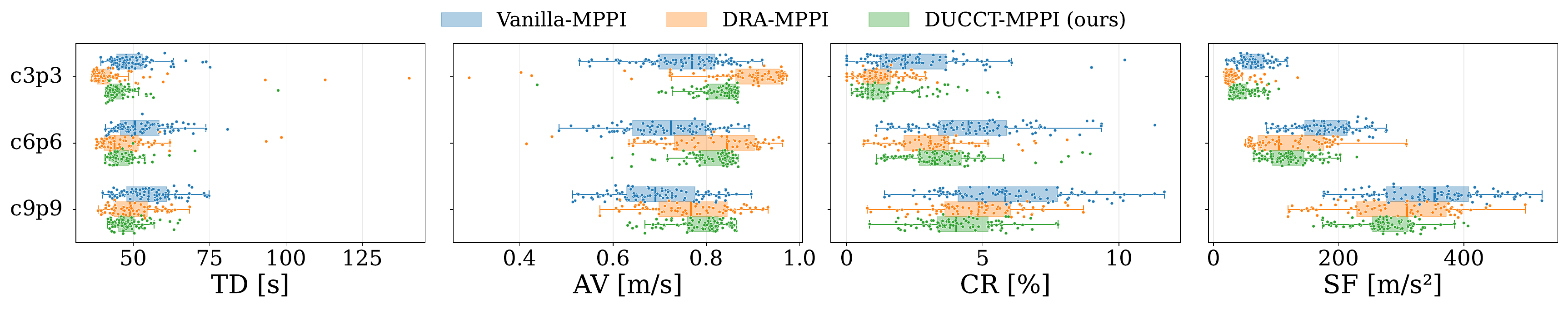}
    \caption{Per-run distributions of travel duration, average velocity, collision rate, and social force are shown as boxplots for all controllers and scenarios.}
    \label{fig:baseline_boxplots}
\end{figure*}

To ensure statistical robustness, we evaluate each experimental configuration over a minimum of 100 runs. We average the per-run metrics using only successful runs and report the mean and standard deviation in \autoref{tab:experiment_metrics}. The distribution of the per-run metrics is shown in \autoref{fig:baseline_boxplots}.

\begin{table*}[htbp]
\centering
\caption{Navigation performance metrics for three controllers (Vanilla-MPPI, DRA-MPPI, and DUCCT-MPPI) across three interaction densities (c3p3, c6p6, c9p9). The table lists mean and standard deviation. Arrows indicate whether lower ($\downarrow$) or higher ($\uparrow$) values are preferable.}
\label{tab:experiment_metrics}
\setlength{\tabcolsep}{5pt}
\renewcommand{\arraystretch}{0.90}
\begin{tabular}{ll rrrrr}
\toprule
Scenario & Controller & TD [s] $\downarrow$ & AV [m/s] $\uparrow$ & CR [\%] $\downarrow$ & SF [m/s²] $\downarrow$ & SR [\%] $\uparrow$ \\
\midrule
\multirow{3}{*}{c3p3} & Vanilla-MPPI & 49.69 (7.62) & 0.76 (0.09) & 2.54 (1.86) & 62.55 (23.39) & 91.18 \\
 & DRA-MPPI & \textbf{43.62 (14.66)} & \textbf{0.88 (0.12)} & \textbf{1.14 (0.81)} & \textbf{32.90 (20.45)} & 86.41 \\
 & DUCCT-MPPI (ours) & 45.02 (6.16) & 0.83 (0.05) & 1.42 (1.18) & 45.38 (17.75) & \textbf{97.17} \\

\midrule

\multirow{3}{*}{c6p6} & Vanilla-MPPI & 52.69 (8.36) & 0.72 (0.09) & 4.70 (2.05) & 177.75 (46.93) & 62.67 \\
 & DRA-MPPI & 47.39 (10.60) & 0.81 (0.11) & \textbf{3.18 (1.51)} & \textbf{122.00 (57.84)} & 75.00 \\
 & DUCCT-MPPI (ours) & \textbf{46.28 (4.75)} & \textbf{0.81 (0.05)} & 3.50 (1.49) & 122.40 (35.15) & \textbf{90.83} \\

\midrule

\multirow{3}{*}{c9p9} & Vanilla-MPPI & 55.08 (8.65) & 0.69 (0.09) & 6.02 (2.30) & 344.54 (86.27) & 47.83 \\
 & DRA-MPPI & 49.74 (7.28) & 0.77 (0.09) & 4.73 (1.76) & 292.61 (91.06) & 50.38 \\
 & DUCCT-MPPI (ours) & \textbf{48.23 (4.97)} & \textbf{0.79 (0.06)} & \textbf{4.20 (1.34)} & \textbf{274.07 (49.91)} & \textbf{77.97} \\
\bottomrule
\end{tabular}
\end{table*}

As detailed in \autoref{tab:experiment_metrics} and \autoref{fig:baseline_boxplots}, DUCCT-MPPI shows superior robustness across all densities. In the sparse setting (c3p3), DRA-MPPI navigates aggressively (lowest TD, highest AV) but sacrifices reliability, recording the lowest success rate (SR). Conversely, DUCCT-MPPI mitigates edge-case failures, achieving a 97.2\% SR while remaining highly efficient, proving the value of principled uncertainty handling even in simple scenarios.

As pedestrian density increases (c6p6, c9p9), Vanilla-MPPI degrades severely in both SR and social compliance. DUCCT-MPPI consistently achieves the highest SRs, outperforming DRA-MPPI by nearly 28\% in the densest scene (c9p9). Crucially, this safety does not induce conservative freezing: DUCCT-MPPI simultaneously achieves the lowest travel times, highest velocities, and lowest social forces in complex settings.

Overall, these results empirically demonstrate that DUCCT-MPPI exhibits superior robustness to increasing interaction complexity, achieving an optimal balance between efficiency, safety, and social compliance. However, aggregate navigation metrics alone do not assess the statistical validity of the internally predicted collision probabilities, motivating the calibration analysis presented next.

\subsection{Calibration evaluation of DUCCT collision probability} \label{subsec:exp_risk_validity}
\subsubsection{Metrics}
To evaluate the accuracy of the risk estimation provided by the DUCCT method, we compare the controller's predicted collision probability against the binary ground truth outcome $C_{t} \in \{0,1\}$ at each time step.
Since MPPI executes only the first command $\mathbf{u}_0$ in a receding-horizon fashion,
we evaluate the predicted collision risk for the immediate next state.
We define the executed-step risk estimate as the importance-weighted average of rollout risks:
\begin{equation}
    \hat{P}_{t} = \sum_{k=1}^{K} w_k \, \hat{\mathcal{P}}_{k,1},
    \label{eq:executed_step_risk}
\end{equation}
where $w_k$ are the MPPI importance weights from \eqref{eq:mppi_update} and
$\hat{\mathcal{P}}_{k,1}$ denotes the rollout-specific joint collision probability at the first prediction step,
computed via \eqref{eq:collision_est}. We evaluate this prediction using:

\begin{itemize}
    \item \textit{Average Predicted Risk (APR)}: The mean value of $\hat{P}_{t}$ over the experiment.
    
    \item \textit{Brier Score (BS)}: A proper scoring rule measuring the mean squared error of the probability predictions, averaged over all $N$ time steps in the experiment.
    \begin{equation}
        BS = \frac{1}{N} \sum_{t=1}^{N} (\hat{P}_{t} - C_{t})^2
    \end{equation}
    
    \item \textit{Log Loss (LL)}: The negative log-likelihood of the predicted collision probabilities.
    \begin{equation}
    \mathrm{LL} = -\frac{1}{N}\sum_{t=1}^{N}
    \big[ C_t \log \hat{P}_t + (1-C_t)\log(1-\hat{P}_t) \big]
    \end{equation}
\end{itemize}
\noindent

\subsubsection{Controlled Uncertainty Miscalibration}
To analyze calibration effects, we manipulate both localization and prediction uncertainty to induce \textit{underconfident} and \textit{overconfident} regimes, using the parameter configurations summarized in \autoref{tab:calib_setups}.
Localization calibration is adjusted by uniformly scaling AMCL motion model parameters $\alpha_1$--$\alpha_5$ (larger $\alpha$: inflated covariance; smaller $\alpha$: underestimated covariance).
Prediction calibration is manipulated by (i) the diagonal initialization covariance $\Sigma_{\text{init}}$ of the pedestrian filter and (ii) a multiplicative covariance inflation factor $\kappa_{\text{pred}}$ applied to predicted positional covariances: $\Sigma_{t+1} = \kappa_{\text{pred}} \Sigma_t$.
DUCCT-MPPI does not internally correct upstream uncertainty. It uses AMCL and predictor covariances as provided, so the miscalibrated conditions explicitly test the robustness of chance-constrained planning under systematically biased uncertainty.

\begin{table}[htbp]
\centering
\renewcommand{\arraystretch}{1.0}
\begin{tabular}{|c|c|cc|}
\hline
\multirow{2}{*}{\textbf{Setup}} 
& \textbf{Localization} 
& \multicolumn{2}{c|}{\textbf{Prediction}} \\ \cline{2-4}
& $\boldsymbol{\alpha_{1\dots5}}$  
& $\boldsymbol{\Sigma_{\text{init}}}$ & $\boldsymbol{\kappa_{\text{pred}}}$ \\ \hline
Standard        & 0.2        & 0.01     & 1.0 \\ \hline
Underconfident  & 0.4        & 0.05     & 1.2 \\ \hline
Overconfident   & $10^{-3}$  & $10^{-4}$& 1.0 \\ \hline
\end{tabular}
\vspace{-0.5ex}
\caption{Localization and prediction calibration setups used in the experiments are shown, specifying the AMCL motion noise parameters $\alpha_{1\dots5}$, the initial prediction covariance $\Sigma_{\text{init}}$, and the prediction covariance scaling factor $\kappa_{\text{pred}}$. Three configurations are evaluated: standard, underconfident, and overconfident.The standard configuration represents the empirically untuned default parameters used for the baseline comparison.}
\label{tab:calib_setups}
\vspace{-1.0ex}
\end{table}

\begin{figure*}[t]
    \centering
    \includegraphics[width=\textwidth]{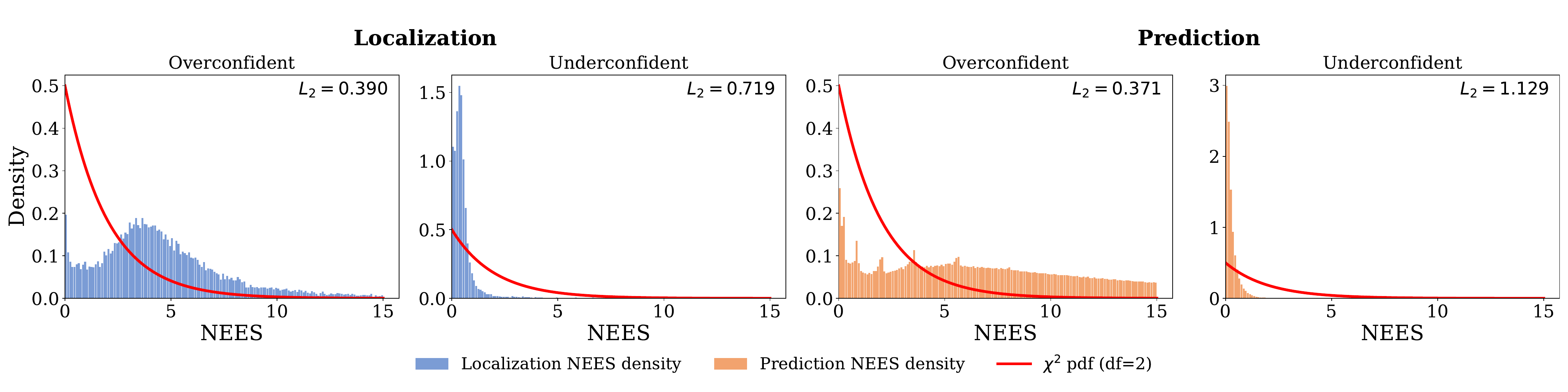}
    \caption{Normalized Estimation Error Squared (NEES) distributions for AMCL localization and pedestrian prediction compared against the theoretical $\chi^2$ distribution with two degrees of freedom. Alignment between empirical NEES density and the reference $\chi^2$ curve indicates statistically consistent uncertainty estimates. A systematic leftward shift (lower NEES values) reflects underconfident covariance estimates, i.e., reported uncertainty exceeds the empirical error. Conversely, a rightward shift (higher NEES values) indicates overconfidence, where the true estimation error is larger than predicted by the covariance.}
    \label{fig:L2_verification}
\end{figure*}

To verify that the intended calibration regimes are realized, we perform consistency checks between reported covariances and empirical errors using NEES-based statistics and an $L_2$ divergence measure between empirical and theoretical distributions, following \cite{Tsuei_2023_Uncertainty}.
We report these verification results for localization and prediction modules in Figure \autoref{fig:L2_verification}. The verification was done with the c9p9 scenario.

\subsubsection{Confidence Sensitivity and Failure Modes} \label{subsec:exp_confidence_ablation}

To understand how miscalibrated uncertainty propagates through the risk-aware planning pipeline, we evaluate DUCCT-MPPI under the four possible combinations of overconfident and underconfident localization ($C_{\text{loc}}$) and prediction ($C_{\text{pred}}$). The results across the three density scenarios are summarized in \autoref{tab:experiment_metrics}.

\begin{table*}[htbp]
\centering
\caption{Confidence ablation results across three pedestrian densities. Arrows ($\uparrow$, $\downarrow$) indicate the optimal direction. Best results per scenario are bolded. Values are mean (std), except SR (mean only).}
\label{tab:experiment_metrics}
\setlength{\tabcolsep}{5pt}
\renewcommand{\arraystretch}{0.90}
\begin{tabular}{lll rrrrrrrr}
\toprule
Scenario & $C_{\text{pred}}$ & $C_{\text{loc}}$ & TD [s] $\downarrow$ & AV [m/s] $\uparrow$ & CR [\%] $\downarrow$ & SF [m/s²] $\downarrow$ & SR [\%] $\uparrow$ & APR [\%] & BS $\downarrow$ & LL $\downarrow$ \\
\midrule
\multirow{4}{*}{c3p3} & \multirow{2}{*}{over} & over & \textbf{42.97 (2.13)} & \textbf{0.85 (0.02)} & 1.39 (1.15) & 42.13 (18.38) & \textbf{99.10} & 0.05 (0.03) & 0.0138 (0.0112) & 0.2755 (0.1843) \\
 &  & under & 43.23 (2.43) & 0.85 (0.03) & 0.98 (0.82) & 37.81 (14.17) & 99.10 & 0.02 (0.01) & 0.0099 (0.0083) & 0.2432 (0.1675) \\
\cmidrule(lr){2-11}

 & \multirow{2}{*}{under} & over & 58.41 (17.53) & 0.71 (0.12) & 1.61 (1.51) & 73.21 (32.18) & 99.10 & 0.89 (0.61) & 0.0131 (0.0111) & 0.0432 (0.0330) \\
 &  & under & 43.17 (2.00) & 0.85 (0.02) & \textbf{0.88 (0.72)} & \textbf{37.63 (15.84)} & 98.04 & 0.32 (0.10) & \textbf{0.0083 (0.0066)} & \textbf{0.0333 (0.0240)} \\

\midrule

\multirow{4}{*}{c6p6} & \multirow{2}{*}{over} & over & 46.44 (9.39) & 0.81 (0.07) & 3.12 (1.15) & \textbf{116.01 (37.67)} & \textbf{99.09} & 0.07 (0.03) & 0.0304 (0.0105) & 0.6728 (0.2506) \\
 &  & under & \textbf{43.96 (2.30)} & 0.83 (0.03) & 3.57 (1.33) & 123.64 (32.59) & 83.74 & 0.04 (0.01) & 0.0355 (0.0136) & 0.7719 (0.2402) \\
\cmidrule(lr){2-11}

 & \multirow{2}{*}{under} & over & 62.58 (16.76) & 0.68 (0.11) & \textbf{2.46 (1.53)} & 180.30 (49.74) & 98.75 & 1.12 (0.47) & \textbf{0.0208 (0.0117)} & \textbf{0.0731 (0.0374)} \\
 &  & under & 44.50 (3.88) & \textbf{0.83 (0.04)} & 3.87 (1.24) & 123.20 (38.00) & 91.35 & 0.67 (0.15) & 0.0348 (0.0110) & 0.1366 (0.0402) \\

\midrule

\multirow{4}{*}{c9p9} & \multirow{2}{*}{over} & over & 47.30 (5.90) & \textbf{0.80 (0.06)} & 4.82 (1.70) & \textbf{275.56 (68.03)} & 87.72 & 0.09 (0.03) & 0.0472 (0.0167) & 1.0414 (0.3412) \\
 &  & under & \textbf{46.65 (4.48)} & \textbf{0.80 (0.05)} & 5.31 (1.68) & 299.11 (73.34) & 47.25 & 0.04 (0.01) & 0.0530 (0.0173) & 1.1614 (0.4007) \\
\cmidrule(lr){2-11}

 & \multirow{2}{*}{under} & over & 56.93 (8.17) & 0.72 (0.07) & \textbf{4.06 (2.15)} & 341.42 (52.45) & \textbf{95.41} & 1.59 (0.59) & \textbf{0.0348 (0.0164)} & \textbf{0.1270 (0.0549)} \\
 &  & under & 47.05 (4.65) & \textbf{0.80 (0.05)} & 4.83 (1.62) & 302.46 (57.02) & 67.33 & 0.76 (0.18) & 0.0448 (0.0154) & 0.1971 (0.0682) \\
\bottomrule
\end{tabular}
\end{table*}

Rather than uniformly degrading performance, miscalibration induces three distinct, highly asymmetric behavioral phenomena:

\paragraph*{The Penalty of Overconfidence} 
When the prediction module is overconfident ($C_{\text{pred}} = \text{over}$), the planner is effectively blind to edge-case interactions. This is evidenced by a severe degradation in scoring rules. For example, in the c9p9 scenario, overconfident prediction configurations yield Log Loss (LL) values exceeding $1.0414$, compared to $0.1270$ for the underconfident counterpart. The model predicts a very low Average Predicted Risk (APR $< 0.1\%$), heavily penalizing the system when inevitable collisions occur. This validates that while overconfident models may satisfy mathematical safety constraints during the planning phase, they routinely violate them during execution.

\paragraph*{The Freezing Robot Phenomenon} 
Conversely, utilizing underconfident prediction paired with overconfident localization ($C_{\text{pred}} = \text{under}, C_{\text{loc}} = \text{over}$) induces extreme conservatism. In this regime, the system achieves exceptionally high Success Rates (SR) across tested scenarios, maintaining a robust $95.41\%$ even in the highly dense c9p9 scenario where other configurations begin to fail. However, this safety comes at a prohibitive cost to efficiency: Travel Duration (TD) spikes significantly (e.g., $62.58\,\mathrm{s}$ in c6p6) and Average Velocity (AV) drops to its lowest levels. The inflated uncertainty generates overly conservative probability estimates (APR jumping to $1.59\%$ in c9p9), demonstrating the classic "freezing robot" problem where the planner struggles to find feasible, confident paths through dynamic obstacles.

\paragraph*{The Paradox of Probability Dilution} 
Intuitively, one might assume that artificially inflating the uncertainty of \textit{both} prediction and localization ($C_{\text{pred}} = \text{under}, C_{\text{loc}} = \text{under}$) would yield the most conservative, and therefore safest, behavior. Our results reveal the exact opposite. In the dense c9p9 scenario, moving from a single underconfident module to dual underconfident modules causes the Success Rate to plummet from $95.41\%$ to $67.33\%$. Furthermore, the robot paradoxically speeds up (AV increases from $0.72\,\mathrm{m/s}$ to $0.80\,\mathrm{m/s}$) while the Average Predicted Risk (APR) drops from $1.59\%$ to $0.76\%$. 

We attribute this counter-intuitive failure mode to \textit{probability dilution}. When the covariance of both the robot's pose and the pedestrians' future trajectories are simultaneously inflated, the probability mass becomes excessively smeared across the spatial costmap. Consequently, the local risk density flattens and falls below the planner's decision threshold $\sigma$. The planner calculates a spuriously low collision risk, encouraging unsafe, aggressive behavior despite the massive global uncertainty.

These ablation results highlight a critical takeaway for risk-aware motion planning: safety cannot be achieved by simply inflating covariance matrices to establish upper bounds on uncertainty. Due to probability dilution, static over-approximations of uncertainty can silently destroy safety constraints. Consequently, chance-constrained frameworks like DUCCT-MPPI fundamentally require dynamically calibrated uncertainty estimates to maintain valid operational guarantees.

\section{Conclusion and Future Work}
\label{sec:conclusion}

In this paper, we presented DUCCT-MPPI, a chance-constrained model predictive control framework that explicitly integrates both localization and pedestrian prediction uncertainties into a unified risk-aware cost formulation. Through extensive simulation in dense social navigation scenarios, we demonstrated that DUCCT-MPPI achieves competitive efficiency while scaling more robustly to complex human-robot interactions compared to baseline MPPI formulations. 

Beyond nominal performance, we evaluated the statistical validity of risk estimates under controlled miscalibration. We demonstrated that while overconfidence leads to safety violations, systematically inflating uncertainty (underconfidence) does not guarantee conservatism. Instead, simultaneously over-approximating both localization and prediction covariance triggers \textit{probability dilution}, a paradoxical state where the spatial risk density flattens below the planner's decision threshold, inducing unsafe behavior despite massive global uncertainty. 

These findings carry important implications for the deployment of probabilistic motion planners. They highlight that establishing static upper bounds on covariance is insufficient for safety; risk-aware planners fundamentally require accurately calibrated, dynamic uncertainty estimates to maintain theoretical constraints during execution.

Future work will focus on integrating online calibration mechanisms, such as Conformal Prediction, to adjust covariance in real-time. Furthermore, while physics-based simulation successfully isolated the effects of miscalibration, further study is required to assess performance under the non-Gaussian noise and computational constraints of embedded hardware; we therefore plan to validate DUCCT-MPPI on physical robot platforms in real-world pedestrian environments.

\bibliographystyle{IEEEtran}
\bibliography{biblio}

\end{document}